# DGCNet: An Efficient 3D-Densenet based on Dynamic Group Convolution for Hyperspectral Remote Sensing Image Classification

Guandong Li[1*]

1.*Suning, Xuanwu, Nanjing, 210042, Jiangsu, China.

*Corresponding author(s). E-mail(s): leeguandon@gmail.com

*Abstract:* Deep neural networks face many problems in the field of hyperspectral image classification, lack of effective utilization of spatial spectral information, gradient disappearance and overfitting as the model depth increases. In order to accelerate the deployment of the model on edge devices with strict latency requirements and limited computing power, we introduce a lightweight model based on the improved 3D-Densenet model and designs DGCNet. It improves the disadvantage of group convolution. Referring to the idea of dynamic network, dynamic group convolution(DGC) is designed on 3d convolution kernel. DGC introduces small feature selectors for each grouping to dynamically decide which part of the input channel to connect based on the activations of all input channels. Multiple groups can capture different and complementary visual and semantic features of input images, allowing convolution neural network(CNN) to learn rich features. 3D convolution extracts high-dimensional and redundant hyperspectral data, and there is also a lot of redundant information between convolution kernels. DGC module allows 3D-Densenet to select channel information with richer semantic features and discard inactive regions. The 3D-CNN passing through the DGC module can be regarded as a pruned network. DGC not only allows 3D-CNN to complete sufficient feature extraction, but also takes into account the requirements of speed and calculation amount. The inference speed and accuracy have been improved, with outstanding performance on the IN, Pavia and KSC datasets, ahead of the mainstream hyperspectral image classification methods.

1. Introduction

Hyperspectral images(HSIs) usually have dozens to hundreds of bands, and are narrow-band imaging that can obtain nearly continuous spectral information, which can provide rich spectral information to enhance the ability to distinguish ground objects[1]. While collecting each sample point, the hyperspectral imager also collects its spatial geometric distribution information, forming a three-dimensional data cube in which two-dimensional spatial information and spectral information are integrated. HSI classification is an essential part of many applications of hyperspectral remote sensing, and is widely used in environmental monitoring, traffic planning, crop yield estimation, land use, and engineering surveying [2].

In recent years, deep learning has become one of the most successful techniques and has achieved impressive performance in the field of computer vision. Motivated by these major breakthroughs, deep learning was introduced to classify HSIs in the remote sensing domain [3, 4, 5, 6]. Compared with traditional methods of manually designing features, deep learning can automatically learn high-level abstract features from complex hyperspectral data. In earlier studies, due to the vector-based input requirement in the network layers, hyperspectral pixels were flattened into one-dimensional feeds into fully connected networks. In SAE [7] or DBN [8], [9], the original spectral vector is directly used for training, however the flattened training samples lose the spatial information of the original image. Liu et al. [10] proposed an efficient classification framework based on deep learning and active learning, in which DBN was used to extract deep spectral features, and an active learning algorithm was employed to iteratively select high-quality labeled samples as training samples. In [11], [12], PCA is first used to reduce the dimensionality of the entire hyperspectral data, and then the spatial information contained in the nearby regions of the input hyperspectral is reduced, and a 2D-CNN is used for feature extraction.The above method combines PCA and CNN, which not only extracts discriminative spatial features, but also reduces the computational cost. In addition, Zhao et al. [13] proposed a Spectral Spatial Feature Based Classification (SSFC) framework for HSI classification. Under this framework, spectral and spatial features are extracted by Balanced Local Discriminant Embedding (BLDE) and CNN, respectively, and then the spectral and spatial features are fused to train a multi-feature-based classifier. Yue et al. [14] used deep CNN to extract deep features, combined with logistic regression classifier for classification. However, methods such as 2D-CNN [15] [16] need to separate spatial and spectral information to extract features, cannot fully utilize the joint spatial and spectral information, and require complex preprocessing. Zhong et al. [17] proposed to use supervised spatial spectral residual network SSRN based on 3D-CNN, and designed continuous spatial and spectral residual modules to extract spatial and spectral information, respectively. Wang [18] et al. proposed a fast and dense spatial spectral convolution network FDSSC, similar to SSRN, constructing 1D-CNN and 3D-CNN dense blocks and concatenating them into a deep network. [19] based on the double convolution pooling structure, directly using stacked 3D-CNN for classification, and achieved good results. Fsknet [20] proposed a 3d to 2d module and a selective kernel mechanism. 3D-SE-Densent [21] introduced the se mechanism in 3dcnn to correlate the convolutional feature maps between different channels, activate valid information in the feature map, and suppress invalid information. Therefore, using 3D-CNN to extract the spatial and spectral information of hyperspectral remote sensing images has become an important direction of hyperspectral remote sensing image classification. We also pay attention to some transformer-based methods [22, 23, 24]. [25, 26] adopted grouped spectral embedding and transformer encoder modules to model spectral representations. However these methods have obvious shortcomings, they all treat spectral bands or spatial patches as tokens and encode all tokens, resulting in a lot of redundant computation, whereas HSI data already contains a lot of redundant information. The accuracy of the transformer-based method is often not as good as that based on 3D-CNN, and the amount of calculation is also greater than that of 3D-CNN.

  3D-CNN has the ability to sample in both spatial and spectral dimensions at the same time, which can ensure the effectiveness of spectral feature extraction while retaining the spatial feature extraction capability of 2d convolution. 3D-CNN has the ability to directly process high-dimensional data, and abandons the method of reducing the dimension of hyperspectral images in advance. However, due to the introduction of the spectral dimension into the convolution kernel, the parameter quantity of the feature map is greatly increased. In particular, the input spectral dimension is high, and the amount of calculation and model parameters will increase exponentially. However, usually in order to reduce the amount of parameters, the use of 3d convolution is reduced in the model, which weakens the feature extraction of convolution kernels. These models are computationally expensive and cannot be deployed on edge devices with strict latency requirements and limited computing resources. And the currently widely used methods such as DFAN[27], MSDN[28], 3D-Densenet[29], 3D-SE-Densenet all use operations such as dense connection. Dense connections directly connect each layer with all layers before it for feature reuse, but dense connections introduce redundancy when earlier features are not needed in subsequent layers. Therefore, how to introduce better lightweight structures and more effective feature extraction capabilities into 3d convolution becomes the direction of faster hyperspectral classification.

Replacing ordinary convolutions with group convolutions can significantly improve the computational efficiency of deep convolutional networks, which has been widely adopted in compact network architecture design. Standard group convolution divides the input and output channels in a convolutional layer equally into G mutually exclusive groups, while performing regular convolution operations within a single group, theoretically reducing the computational burden by a factor of G. However, the existing group convolution has two major disadvantages.1. Due to the introduction of sparse connections, the expressive ability of convolution is weakened, resulting in a decrease in performance for difficult samples; 2. A fixed connection pattern that does not change based on the characteristics of the input samples. Refer to the idea of dynamic networks [30][31][32], driven by channel or pixel-level methods. This paper introduces dynamic grouped convolution on a 3d convolution kernel, introducing a small feature selector for each grouping to dynamically decide which part of the input channel to connect based on the activations of all input channels. Multiple groups can capture different complementary visual and semantic features of input images, allowing dgc to learn rich features. Due to the sparse nature of hyperspectral data, the spatial distribution of samples is uneven, and there is a lot of redundant information in the spectral dimension. 3D convolution extracts high-dimensional and redundant hyperspectral data, and there is also a lot of redundant information between convolution kernels. DGC module allows 3D-CNN to select channel information with richer semantic features and discard inactive regions. It not only allows 3D-CNN to complete sufficient feature extraction, but also takes into account the requirements of speed and calculation amount, and is compatible with existing 3D-CNN to ensure the complete structure of the original network. For inference, zero-value pruning is performed on the weights in the DGC module. At this time, the DGCNet can be regarded as a pruned network.

The main contributions of this paper are as follows:

1. In this paper, an improved and efficient 3D-Densenet based spatial spectral hyperspectral image classification method based on the DGC module is proposed. Using 3D-CNN as the basic structure, combined with the lightweight convolution design, the dense connection is combined with the DGC module, the dense connection is conducive to the feature reuse in the network, and the 3D-Densenet model of the combined DGC is designed.

2. This paper introduces a lightweight structure design in 3D-CNN, DGC improves the drawbacks of the existing group convolution, introduces a small feature selector for each group, and dynamically selects features. The pruned model is used at the time of inference, which greatly reduces the amount of parameters and can be applied to the scene of edge devices.

3. Compared with the network combining various mechanisms, DGCNet is more concise, without complicated connections, and the amount of calculation is less. We divided DGCNet into three models of small/base/large. DGCNet has achieved the leading structure in the three data sets of IN, Pavia, and KSC, and is still competitive under the condition of lightweight design.

2. Spatial spectral classification method based on lightweight convolutional structure

2.1 Efficient architecture deisgn

As special cases of sparsely connected convolution, grouped convolution and depthwise separable convolution are the most commonly used modules in lightweight structure design. Depthwise separable convolution can be seen as an extreme case of group convolution, one conv kernel corresponds to one channel dimension, as shown in Fig1. AlexNet first used group convolutions to deal with memory constraints. ResNeXt further applied group convolution and demonstrated its effectiveness. CondenseNet and FLGC automatically learn the connections of group convolutions during training. All these existing group convolutions have fixed connections in the inference process, which inevitably weakens the representational power of convolutions due to the sparse structure. Our proposed DGC can effectively solve this dilemma by adopting a dynamic execution strategy, which can maintain sparse computation without destroying the original network structure. Group convolution is a grouping feature extraction in the channel dimension. The spectral sparse and redundant characteristics of hyperspectral images are very suitable for the group convolution mode. We did not use separable convolution because it cannot take full advantage of the redundant features of hyperspectral information, which is a single-channel model.

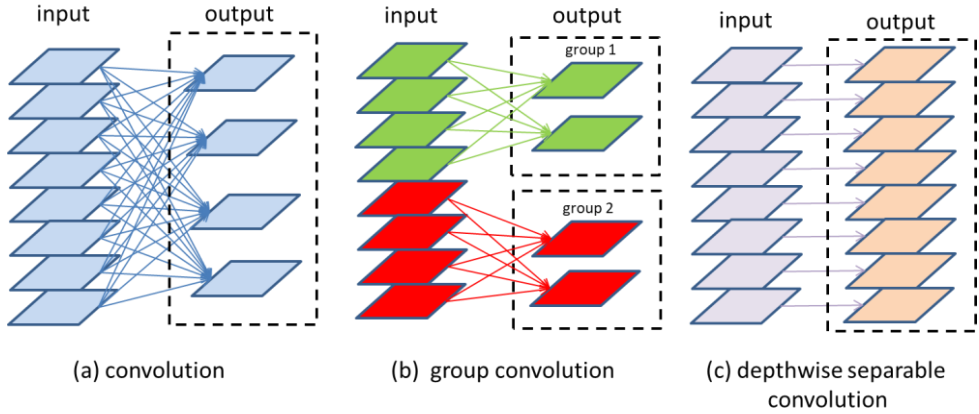

Fig1: Schematic diagram of convolution, group cpnvolution and depthwise separable convolution

Convolutional network inference efficiency can be improved by weight pruning [33, 34] or weight quantization [35, 36]. These methods are efficient because deep networks often have a large number of redundant weights, which can be pruned or quantized without sacrificing accuracy. For convolutional networks, different pruning techniques may result in different levels of granularity. Such as independent weight pruning [37], such fine-grained pruning can often achieve a high degree of sparsity. DGCNet can also be regarded as a dynamic pruning network, which can generate efficient group convolution and achieve the optimal point between sparsity and regularity. The input is MXxNxLxC, where MxN is the spatial dimension, L is the spectral dimension, and C is the channel dimension. The pruning function of DGC is shown in Fig 2. The DGC module is used during training, but the zero-valued weights in the DGC weights are deleted during inference. At this time, the network is a devalued model, and the amount of parameters and calculations are greatly reduced.

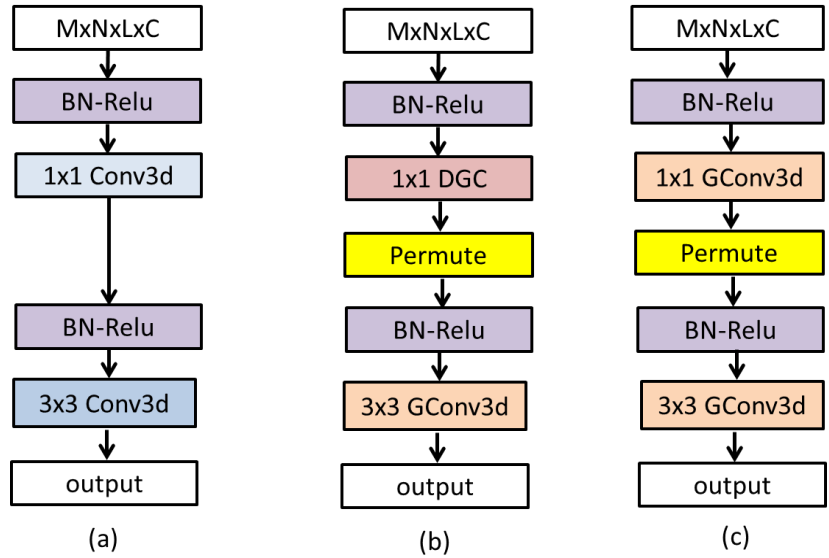

Fig2: (a) is the feature extraction cnn module in 3D-Densenet, (b) is the DGC module during training, (c) is the DGC module during testing.

2.2 Dynamic group convolution

There is a lack of sample data for hyperspectral image, and the sample features are sparse, unevenly distributed in spatial, and there is a large amount of high-dimensional redundant information in the spectral dimension. Although the use of the 3D-CNN structure can utilize the spatial spectral information, how to more effectively realize the depth extraction of the spatial spectral information is still a problem worthy of attention. As the core of the CNN, the convolution kernel is usually regarded as an information aggregate that aggregates spatial information and feature dimension information on the local receptive field. CNN consist of a series of convolutional layers, nonlinear layers, and downsampling layers, so that they can capture image features from the global receptive field for description. However, it is difficult to learn a network with strong performance, and there are many works to improve the performance of the network from the spatial dimension. For example, multi-scale information is embedded in the Inception structure, and features on different receptive fields are aggregated to obtain performance gains. On the Residual structure, the depth extraction of the network is realized by fusing the features generated by different blocks. However, such a network has a high training cost, a slow convergence speed, and is prone to overfitting on small sample datasets. 3D-CNN also has a large number of redundant weights in the convolution in feature extraction, and these weights hardly contribute to the final output. We introduced a lightweight convolutional structure and dynamic network mechanism in hyperspectral remote sensing classification, and introduced the DGC module in the improved and efficient 3D-Densenet. The structure of DGC is shown in Fig3. DGC divides the input channels into multiple groups, each group is equipped with auxiliary heads, and each auxiliary head is equipped with an input channel selector to select the channel dimension for convolution operation. The logic of each group is as follows:1. saliency generator to generate importance scores for input dimensions;2. The input channel selector adopts the gating strategy to determine the most important part of the input dimension according to the importance score.3. Do a normal convolution operation on the selected input dimension. Finally, the outputs of all heads are cascaded and shuffled to the subsequent BN layers and activation layers. The DGC module is similar to model pruning. Combined with lightweight group convolution, it can greatly reduce the amount of model parameters and computation. Specifically, it is to automatically obtain the importance of each feature channel through learning, and then use this importance to enhance useful features and suppress features that are not useful for the current task. It is very effective on hyperspectral sparse ground objects. Spectral data is relatively scarce and belongs to the category of small samples. There are large differences in the feature maps obtained by different convolution kernels. Through DGC, combined with the depth characteristics of 3D-Densenet, features can be extracted more effectively.

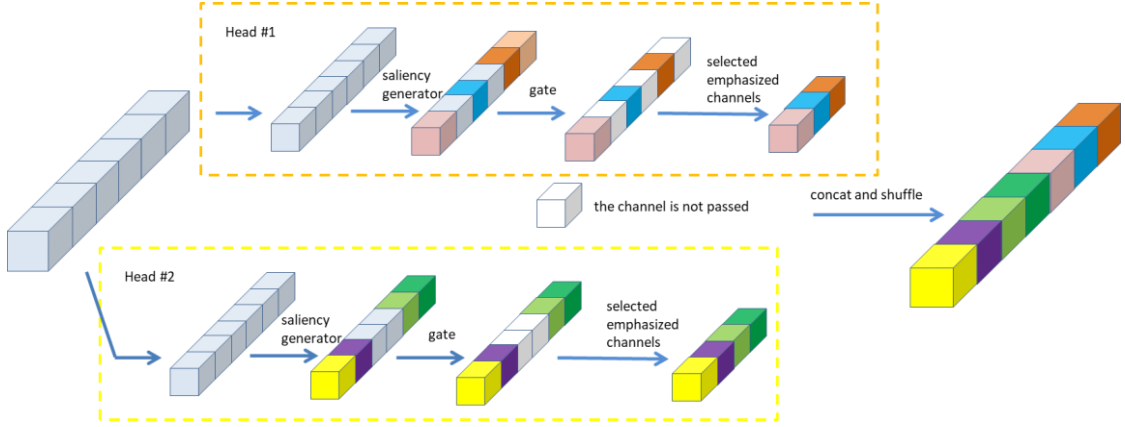

Fig3: Schematic diagram of the DGC layer. The figure shows two sets of heads, the number of channel dimensions of each head is halved compared to the input dimension, and the white blocks represent dimensions that do not participate in convolution elements.

2.2.1 Saliency generator

The saliency generator assigns a score to each input dimension to represent its importance. Each head has a specific saliency generator, which is used to guide different heads to use different input dimensions, thereby increasing the diverse expression of features. The saliency generator follows the design of the SE block. The input is dimensionally reduced by avgpool, and the global spatial spectral information is compressed into a channel descriptor. For the ith head, the importance vector gi is calculated as:

$$g = \left(W(p(x)) + \beta\right)_+$$

g represents the importance vector of the input dimension, $(z)_+$ represents the ReLU activation, and p reduces each input feature map to a single scalar, which uses global average pooling. β and W are learnable parameters, β is the bias, and W is a two-step transformation operation. In order to limit the model complexity and help generalization, two fc layers model the correlation between channels. First, use the compression rate r to reduce the feature dimension in the first fc layer. After one fc, it will rise back to the original dimension in the second fc. x is all input dimensions, and in each head, all input dimensions are candidate vectors.

2.2.2 Gating strategy

After obtaining the importance vector, the next step is to decide which input dimensions are selected by the current head to participate in subsequent convolution operations. To use a threshold to filter input features with lower scores, given the target clipping rate ε, the threshold γ of the ith head satisfies:

$$\varepsilon = \frac{|\{g | g < \gamma, g \in \boldsymbol{g}\}|}{|\{g | g \in \boldsymbol{g}\}|}$$

The importance score serves two purposes. 1.Importance scores less than the threshold will be removed. 2.The remaining dimensions are weighted using the corresponding importance scores to obtain weighted features. Assuming that the number of heads is T and the convolution kernel of the ith head is w, the corresponding convolution calculation is:

$$\boldsymbol{v} = top[(1-\varepsilon)C](\boldsymbol{g})$$
$$\boldsymbol{x}' = \boldsymbol{x}[\boldsymbol{v},:,:]$$
$$\boldsymbol{x}'' = f(\boldsymbol{x}' * \boldsymbol{g}, \boldsymbol{w}) = f(\boldsymbol{y}, \boldsymbol{w}) = \mathrm{y} \times \mathrm{w}$$

It is to select the selected features and corresponding weights for conventional convolution calculation. top[k](z) returns the indices of the largest k elements in z. x is a regular convolution. At the end of the DGC, the outputs are merged and then shuffled to output x'. Fig 4 shows the comparison of SE module and DGC module.

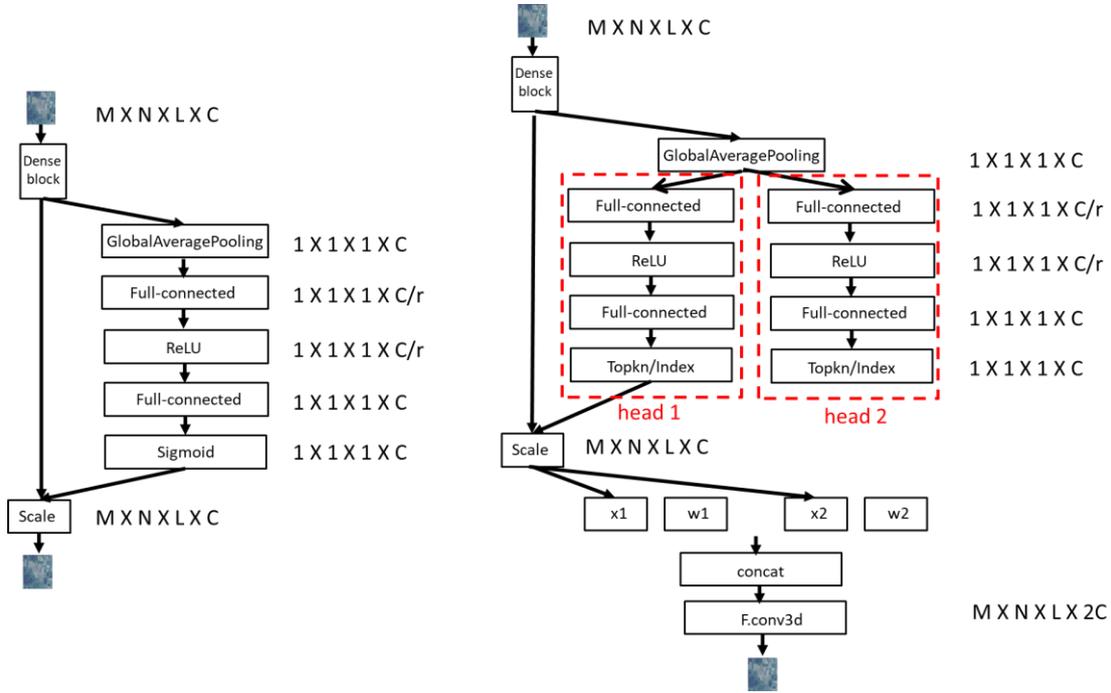

Fig 4    the left is the se module, the right is the dgc module, we chose the head as 2.

2.2.3 computation cost

The regular convolutional MAC with kernel size k is $k^2 C'CH'W'$, In DGC, the MAC of the saliency generator and convolution of each head are $\frac{2C^2}{d}$ and $k^2(1-\varepsilon)C\frac{C'}{T}H'W'$, Therefore, the saving ratio of the MAC of the DGC layer relative to the conventional convolution is:

$$(k^2 C'CH'W'/T(k^2(1-\varepsilon)C\frac{C'}{T}H'W' + \frac{2C^2}{d})) = 1/((1-\varepsilon) + \frac{2TC}{dk^2 C'H'W'}) \approx 1/(1-\varepsilon)$$

The number T of heads has almost no effect on the overall computational cost.

2.3 DGCNet Feature Extraction Framework and Model Implementation
2.3.1 Training DGC network

In the back-propagation stage of the DGC network, only the gradient of the channel dimension weights selected during inference is calculated, and the others are set to zero. In order to prevent the training loss from changing too much due to pruning, the pruning ratio is gradually increased during the training process. The overall training is divided into 3 stages. The first stage (the first 1/12 epochs) is used for warm up, the second stage is to gradually increase the clipping ratio for training, and the third stage (the last 1/4 epochs) is used for the fine-tune sparse network .

2.3.2 DGCNet architecture design

We make two changes to the original 3D-Densenet, which aim to further simplify the architecture and improve its computational efficiency.

**Exponentially increasing with growth rate.** The original Densenet design adds k new feature maps at each layer, where k is a constant called the growth rate. As shown in [38], deeper layers in Densenet tend to rely more on high-level features than low-level features. This motivates us to improve the network by strengthening short connections. We found that this can be achieved by gradually increasing the growth rate with increasing depth. It increases the proportion of features from later layers relative to features from earlier layers. For simplicity, we set the growth rate as $k = 2^{m-1}k_0$, where m is the index of the dense block and $k_0$ is a constant. This way of setting the growth rate does not introduce any additional hyperparameters. The increasing growth rate strategy places a larger proportion of the parameters in later layers of the model. It greatly improves computational efficiency, but may reduce parameter efficiency in some cases.

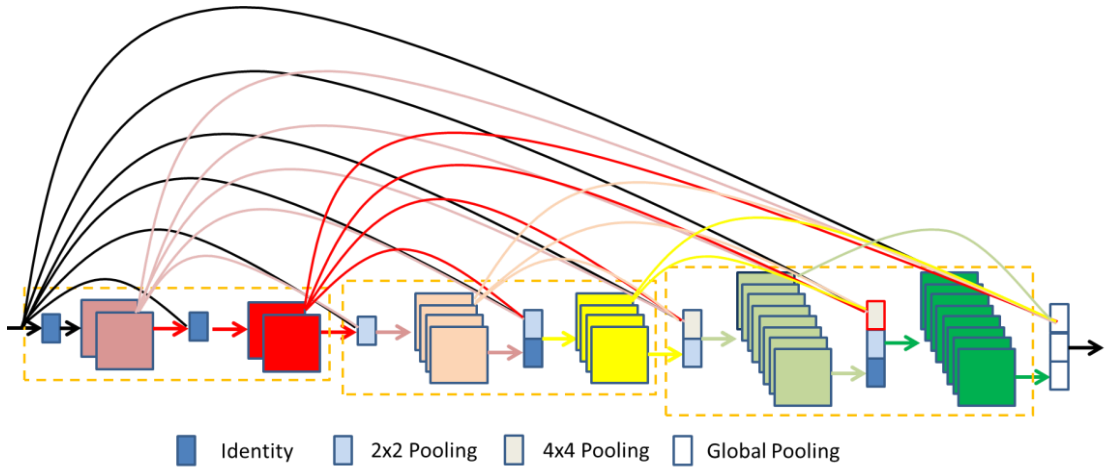

Fig 5. Efficient DenseNet. It differs from the original Densenet in two points: (1) the layers of feature maps of different resolutions are also directly connected; (2) the growth rate doubles as the feature map size shrinks (much more features are generated in the last green dense block than in the first block).

**Fully dense connectivity**. To encourage more feature reuse than the original Densenet architecture, we concatenate the input layer features to all subsequent layers in the network, even if the layers are in different dense blocks (see Fig 5). Since dense blocks have different feature resolutions, we use average pooling to downsample feature maps with higher resolutions when using them as input to lower resolution layers.

The overall model structure of DGCNet is shown Fig6:

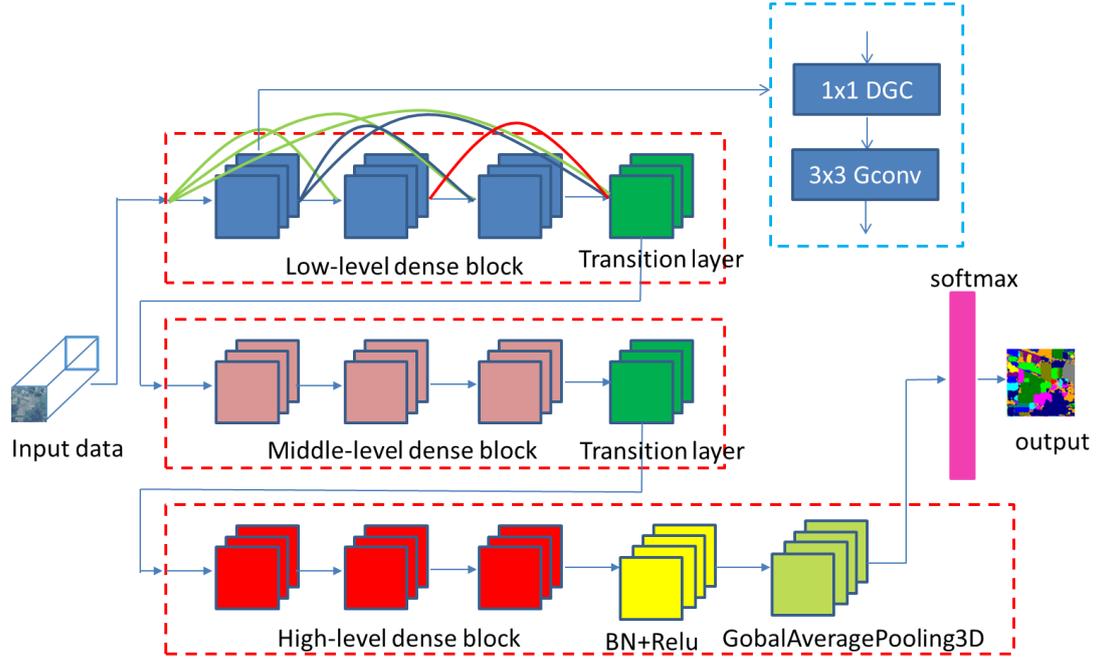

We chose a smaller model to show its structure, which includes 3 stages, the number of dense blocks in the stage is 4, 6, 8, the growth rate is 8, 16, 32, and the head is 2. The model parameter diagram will be shown in the appendix.

3. Experiments and Analysis

In order to evaluate the effect of the DGCNet model, this paper introduces the three most representative hyperspectral image datasets, Indian pines, Pavia University and KSC. The classification metrics used overall classification accuracy (OA), average classification accuracy (AA) and kappa coefficient.

Our proposed DGCNet updates the parameters of the 3D convolution by the gradient of the back propagation loss function. The gradient descent algorithm of the model adopts Adam optimizer[34] to optimize loss function. In all datasets, we trained 200 epochs. The model framework performs a total of four trainings and eventually integrates the classification results on the four best models. During the training process, the model with the highest classification accuracy performance on the validation set is retained every time. The learning rate is 0.0005.

3.1 Experimental dataset

3.1.1 Indian pines dataset

The Indiana Pines dataset was collected in June 1992 by the AVIRIS Spectral Imager at the Indiana Pine Forest Experimental Area in northwestern Indiana, USA. The size of the data image is 145x145 pixels, the spatial resolution is 20m, and there are 220 bands in the wavelength range of 0.4-2.5μm, we removed 20 water vapor absorption and low signal-to-noise ratio bands, and kept the remaining 200 bands for experiments. The dataset includes 16 types of ground objects including grassland, buildings and crops. The dataset includes 16 types of ground objects including grassland, buildings and crops. Its sample spatial distribution is shown in Fig 7.

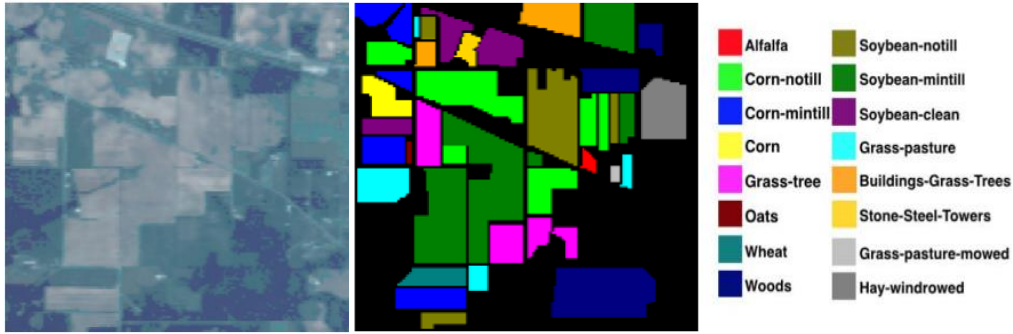

Fig 7 False color image and Ground-truth labels of Indian Pines

### 3.1.2 Pavia University dataset

The Pavia University dataset was collected in 2001 by the ROSIS Spectral Imager in the Pavia region of northern Italy. The image size is 610x340 pixels, the spatial resolution is 1.3m, and there are 115 bands in the wavelength range of 0.43-0.86μm. we removed 12 bands containing strong noise and water vapor absorption, and retained the remaining 103 bands for experiments. The dataset includes 9 ground objects including roads, trees and roofs, and the spatial distribution of different categories is shown in Fig 8.

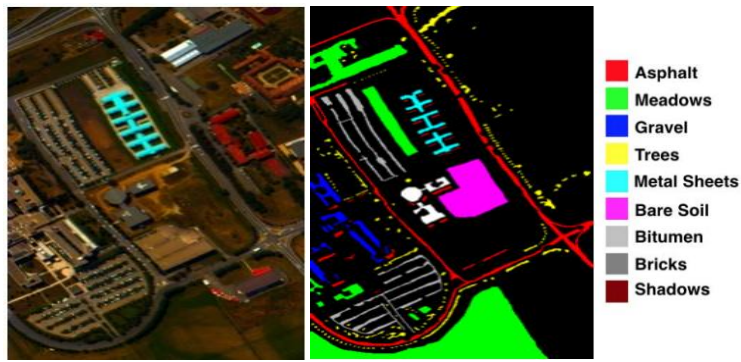

Fig8 False color image and Ground-truth labels of Pavia University dataset

### 3.1.3 KSC dataset

The KSC dataset was collected on March 23, 1996 by the AVIRIS Spectral Imager at Kennedy Space Center, Florida. AVIRIS acquired 224 bands of 10 nm width, centered at 400-2500 nm. The spatial resolution of KSC data acquired from an altitude of approximately 20 km is 18 m. After removing water absorption and low SNR bands, 176 bands were used for analysis. It defines 13 categories.

### 3.2 Experiment analysis

The experimental analysis adopts the structure of 3 stages. The number of dense blocks in each stage is 14, the growth rate is 8, 16, 32, the head is 4, the number of groups in the 3x3 group is 4, the gate_factor is 0.25, and the compression number is 16.

### 3.2.1 Influence of different ratios of training datasets in the DGCNet

The number of training sets is relatively sensitive in hyperspectral data with a small sample size, so the performance of the model under different training, validation and testing scales is discussed. DGCNet selects a ratio of 5:1:4 on both pavia university and ksc datasets, and selects a ratio of 6:1:3 on the indian pines dataset. We noticed that with the increase of the training set, the accuracy of the model is getting higher and higher, but there is a certain threshold for the increase of the training set. After reaching the threshold, if the proportion is increased, the accuracy will not increase significantly and may decline. The neighboring pixel block size is 11.

Table 1 OA,AA and kappa under different training dataset ratios for the Indian Pines dataset ($M \times N=11 \times 11$)

| Ratios | OA | AA | Kappa |
| --- | --- | --- | --- |
| 2:1:7 | 93.68 | 94.32 | 92.79 |
| 3:1:6 | 96.55 | 95.53 | 96.06 |
| 4:1:5 | 98.08 | 98.43 | 97.81 |
| 5:1:4 | 97.78 | 97.83 | 97.48 |
| 6:1:3 | 99.43 | 99.19 | 99.35 |

Table2 OA,AA and kappa under different training dataset ratios for the Pavia University dataset ($M \times N=11 \times 11$)

| Ratios | OA | AA | Kappa |
| --- | --- | --- | --- |
| 2:1:7 | 99.88 | 99.71 | 99.83 |
| 3:1:6 | 99.89 | 99.49 | 99.84 |
| 4:1:5 | 99.73 | 99.02 | 99.63 |
| 5:1:4 | 99.99 | 99.98 | 99.99 |
| 6:1:3 | 99.99 | 99.99 | 99.99 |

Table 3 OA,AA and kappa under different training dataset ratios for the KSC dataset(MxN=11x11)

| Ratios | OA | AA | Kappa |
| --- | --- | --- | --- |
| 2:1:7 | 97.67 | 96.99 | 97.40 |
| 3:1:6 | 99.33 | 99.02 | 99.25 |
| 4:1:5 | 99.50 | 99.39 | 99.44 |
| 5:1:4 | 99.57 | 99.39 | 99.52 |
| 6:1:3 | 99.29 | 99.05 | 99.21 |

3.2.2 Influence of neighboring pixel block size in the DGCNet

The network fills the edge of the input 145x145x200 image into a 155x155x200 image (taking Indian Pines as an example), and selects an MxNxL neighboring pixel block on the 155x155x200 image in turn, where MxN is the spatial sampling size, and L is the spectral dimension. The original image is too large, which is not conducive to the full feature extraction of convolution, the running speed is slow, and the short-term memory usage increases, which requires high hardware platform, so the processing of input neighboring pixel blocks is adopted. The size of neighboring pixel blocks is an important hyperparameter, but the range of neighboring pixel blocks should not be too small, which will easily lead to insufficient receptive field for convolution kernel feature extraction, and no good results locally. On the Indian Pines dataset, the neighboring pixel block size is from 7 to 17, and there is a significant improvement in accuracy.This is also evident on the Pavia University dataset. As the range of neighboring pixel blocks increases, the increase in overall accuracy becomes smaller and smaller, and there is an obvious threshold effect.

Table 4 OA,AA and kappa for different neighboring pixel block sizes for the Indian Pines dataset

| Neighboring pixel block sizes (M=N) | OA | AA | Kappa |
|---|---|---|---|
| 7 | 98.08 | 96.82 | 97.81 |
| 9 | 99.21 | 98.98 | 99.10 |
| 11 | 99.43 | 99.19 | 99.35 |
| 13 | 99.69 | 99.63 | 99.65 |
| 15 | 99.41 | 99.49 | 99.33 |
| 17 | 99.45 | 99.31 | 99.38 |

Table 6 OA,AA and kappa for different neighboring pixel block sizes for the Pavia University dataset

| Neighboring pixel block sizes (M=N) | OA | AA | Kappa |
|---|---|---|---|
| 7 | 99.97 | 99.94 | 99.96 |
| 9 | 99.98 | 99.94 | 99.97 |
| 11 | 99.99 | 99.98 | 99.99 |
| 13 | 99.99 | 99.99 | 99.99 |
| 15 | 99.99 | 99.99 | 99.99 |
| 17 | 99.99 | 99.99 | 99.99 |

3.2.3 model parameters

We divided dgcnet into three types: small, base and larger. oa, aa and kappa are the results tested on the in dataset.

| | stages/dense block | growth rate | OA | AA | Kappa |
|---|---|---|---|---|---|
| Dgcnet-small | 4,6,8 | 8,16,32 | 99.43 | 97.99 | 99.35 |
| Dgcnet-base | 10,10,10 | 8,16,32 | 99.58 | 99.58 | 99.58 |
| Dgcnet-larger | 14,14,14 | 8,16,32 | 99.69 | 99.69 | 99.69 |

Test the resulting params on the Indian pines dataset, where the spectral dimension of in is 200.

| | 3D-CNN | 3D-DenseNet | HybridSN | Dgcnet-small | Dgcnet-base | Dgcnet-larger |
|---|---|---|---|---|---|---|
| Params | 16394652 | 2562452 | 5503108 | 746844 | 1546236 | 3198492 |

3.3 Experimental results and analysis

On the Indian pines dataset, the input size of DGCNet is 13x13x200, on the Pavia University dataset, the input size of DGCNet is 11x11x103, and on the ksc dataset, the input size of DGCNet is 17x17x176. We also evaluated dgcnet-small/base/larger, and selected SAE, 3D-CNN, 3d-densenet, 3D-SE-DenseNet-BC, spectralformer for comparison.

Table 8 Comparison of the classification accuracies (%) of different methods for the Indian Pines dataset

| N | SAE | SSRN | 3D-CNN | 3D-DenseNet($c=12$; | 3D-SE-DenseNet | Spectralformer | Dgcnet-small | Dgcnet-base | Dgcnet-larger |
|---|---|---|---|---|---|---|---|---|---|

|   |   |   |   | k=32) | (c=3) |   |   |   |   |
|---|---|---|---|---|---|---|---|---|---|
| 1 | 81.82 | 100 | 96.88 | 100 | 95.87 | 70.52 | 100 | 100 | 100 |
| 2 | 82.16 | 99.85 | 98.02 | 99.40 | 98.82 | 81.89 | 99 | 99.47 | 99.65 |
| 3 | 77.54 | 99.83 | 97.74 | 99.48 | 99.12 | 91.30 | 99.11 | 99.51 | 100 |
| 4 | 68.11 | 100 | 96.89 | 100 | 94.83 | 95.53 | 97.92 | 97.65 | 97.96 |
| 5 | 94.36 | 99.78 | 99.12 | 100 | 99.86 | 85.51 | 99.49 | 100 | 100 |
| 6 | 94.45 | 99.81 | 99.41 | 100 | 99.33 | 99.32 | 100 | 99.88 | 99.78 |
| 7 | 94.70 | 100 | 88.89 | 100 | 97.37 | 81.81 | 80.57 | 100 | 100 |
| 8 | 94.36 | 100 | 100 | 100 | 100 | 75.48 | 100 | 100 | 100 |
| 9 | 82.56 | 0 | 100 | 100 | 100 | 73.76 | 95.83 | 100 | 100 |
| 10 | 81.28 | 100 | 100 | 99.85 | 99.48 | 98.77 | 99.74 | 98.85 | 99.66 |
| 11 | 84.47 | 99.62 | 99.33 | 99.53 | 98.95 | 93.17 | 99.73 | 99.72 | 99.66 |
| 12 | 83.77 | 99.17 | 97.67 | 98.58 | 95.75 | 78.48 | 99.86 | 99.56 | 99.41 |
| 13 | 96.42 | 100 | 99.64 | 100 | 99.28 | 100 | 99.21 | 100 | 100 |
| 14 | 92.27 | 98.87 | 99.65 | 99.89 | 99.55 | 79.49 | 99.94 | 99.87 | 99.75 |
| 15 | 80.63 | 100 | 96.34 | 99.64 | 98.70 | 100 | 100 | 100 | 100 |
| 16 | 81.82 | 98.51 | 97.92 | 97.10 | 96.51 | 100 | 97.41 | 98.30 | 98.21 |
| OA | 85.47±0.58 | 99.62±0.00 | 98.23±0.12 | 99.85±0.04 | 98.84±0.18 | 81.76 | 99.43 | 99.58 | 99.69 |
| AA | 86.31±1.14 | 93.46±0.50 | 98.80±0.11 | 99.71±0.25 | 98.42±0.56 | 87.81 | 97.99 | 99.55 | 99.63 |
| K | 83.42±0.66 | 99.57±0.00 | 97.96±0.53 | 99.64±0.04 | 98.60±0.16 | 79.19 | 99.35 | 99.53 | 99.65 |

Table 9 Classification accuracy(%) of different methods for the Pavia University dataset

| No | SAE | 3D-CNN | SSRN | 3D-DenseNet (c=3; k=32) | 3D-SE-DenseNet (c=6) | Spectralformer | Dgcnet-larger |
|---|---|---|---|---|---|---|---|
| 1 | 87.24 | 98.56 | 99.96 | 99.81 | 99.32 | 82.73 | 99.99 |
| 2 | 89.93 | 99.77 | 99.99 | 99.99 | 99.87 | 94.03 | 99.96 |
| 3 | 86.48 | 99.05 | 99.64 | 99.95 | 96.76 | 73.66 | 100 |
| 4 | 99.95 | 99.98 | 99.83 | 99.87 | 99.23 | 93.75 | 100 |
| 5 | 95.78 | 99.90 | 99.81 | 100 | 99.64 | 99.28 | 100 |
| 6 | 97.69 | 99.03 | 99.98 | 100 | 99.80 | 90.75 | 100 |
| 7 | 95.44 | 99.71 | 97.97 | 100 | 99.47 | 87.56 | 100 |
| 8 | 84.40 | 97.53 | 99.56 | 99.86 | 99.32 | 95.81 | 99.99 |
| 9 | 100 | 99.86 | 100 | 100 | 100 | 94.21 | 100 |
| OA | 90.58±0.18 | 98.47±0.41 | 99.79±0.01 | 99.94±0.002 | 99.48±0.02 | 91.07 | 99.99 |
| AA | 92.99±0.39 | 99.28±0.31 | 99.75±0.15 | 99.95±0.003 | 99.16±0.37 | 90.20 | 99.98 |
| K | 87.21±0.25 | 98.97±0.21 | 99.87±0.27 | 99.95±0.002 | 99.31±0.03 | 88.05 | 99.99 |

4.Conclusion

We introduce dynamic grouped convolutions on 3d convolution kernels. DGC introduces small feature selectors for each grouping, dynamically deciding which part of the input channel to connect based on the activations of all input channels. Multiple groups can capture different complementary visual and semantic features of input images, allowing DGC to learn rich features. The sparse nature of hyperspectral data, the uneven distribution of samples in spatial, and the existence of a lot of redundant information in the spectral dimension. 3D convolution extracts high-dimensional and redundant hyperspectral data, and there is also a lot of redundant information between convolution kernels. The DGC module allows 3D-CNN to select channel information with richer semantic features and discard inactive regions. It not only allows 3D-CNN to complete sufficient feature extraction, but also takes into account the requirements of speed and calculation amount, and is compatible with existing 3D-CNN to ensure the complete structure of the original network. In forward reasoning, the 3D-CNN passing through the DGC module can be regarded as a pruned network. DGCNet has achieved good results on Indian pines, pavia university and ksc datasets.